\documentclass[times,twocolumn,final,numbers]{elsarticle}

\usepackage{framed,multirow}

\usepackage{amssymb}
\usepackage{latexsym}

\usepackage{url}
\usepackage{xcolor}
\definecolor{newcolor}{rgb}{.8,.349,.1}

\usepackage{times}
\usepackage{epsfig}
\usepackage{graphicx}
\usepackage{amsmath}
\usepackage{amssymb}
\usepackage{float}
\usepackage{subcaption}
\usepackage{epigraph}
\usepackage{hyperref}

\begin{document}

\thispagestyle{empty}

\clearpage
\thispagestyle{empty}
\ifpreprint
  \vspace*{-1pc}
\fi



\clearpage
\thispagestyle{empty}

\ifpreprint
  \vspace*{-1pc}
\else
\fi

 
 
  
 

\clearpage

\ifpreprint
  \setcounter{page}{1}
\else
  \setcounter{page}{1}
\fi

\begin{frontmatter}

\title{Are you from North or South India? A hard race classification task reveals systematic representational differences between humans and machines}

\author[1]{Harish Katti} 
\author[1]{S.P. Arun} 
\address[1]{Centre for Neuroscience, Indian Institute of Science, Bangalore, India}


\begin{abstract}
We make a rich variety of judgments on faces but the underlying features are poorly understood. While coarse categories such as race or gender are appealing to study, they produce large changes across many features, making it difficult to identify the underlying features used by humans. Moreover, the high accuracy of both humans and machines on these tasks rules out any systematic error analysis. Here we propose, demonstrate and benchmark a novel dataset for understanding human face recognition that overcomes these limitations. The dataset consists of 1647 diverse faces from India labeled with their fine-grained race (North vs South India) as well as classification performance of 129 human subjects on these faces. Our main finding is that, while many machine algorithms achieved an overall performance comparable to humans (64\%), their error patterns across faces were qualitatively different and remained so even when explicitly trained to predict human performance. To elucidate the features used by humans, we trained linear classifiers on overcomplete sets of features derived from each face part. This indicated that mouth shape to be the most discriminative part compared to eyes, nose or the external contour. To confirm this prediction, we performed an additional behavioral experiment on humans by occluding various shape parts. Occluding the mouth impaired race classification in humans the most compared to occluding any other face part. Taken together, our results show that studying hard classification tasks can lead to useful insights into both machine and human vision.
\end{abstract}


\end{frontmatter}





\vspace*{-0.3pc}

\section{Introduction}

\epigraph{Just realize where you come from:\\this is the essence of wisdom.}{\textit{Tao Te Ching}, v. 14}

\vspace*{-0.3pc}

Humans make a rich variety of judgments on faces ranging from gender, race, personality, emotional state etc. Understanding the underlying features can enable a variety of AI applications with human-like performance. Coarse race (Caucasian/Black/Asian) \cite{Brooks2010,Fu2014} as well as gender has been studied in computer vision \textcolor{black}{\cite{Tariq2009,Wang2016,Fu2014}}. Coarse categories are an important first step but do not sufficiently constrain the underlying features used by humans for two reasons. First, these categories involve changes in many features, consequently poorly constraining the features. Second, both humans and machines show high accuracy on these tasks, making any systematic error analysis difficult since errors are typically too few in number. Both limitations can be addressed using classification problems with subtle feature variations that are hard for both humans and machines. 

A natural choice then is finer grained face classification. While there has been some work on discriminating between finer grained race such as Chinese/Japanese/Korean \cite{Wang2016}, Chinese sub-ethnicities \cite{Duan2010} and Myanmar \cite{Tin2011}, these studies have not systematically characterized human performance. In fact it is an open question whether and how well humans can discriminate finer grained race across various world populations. 

\begin{figure*}[t]
\begin{center}
\includegraphics[width=\linewidth]{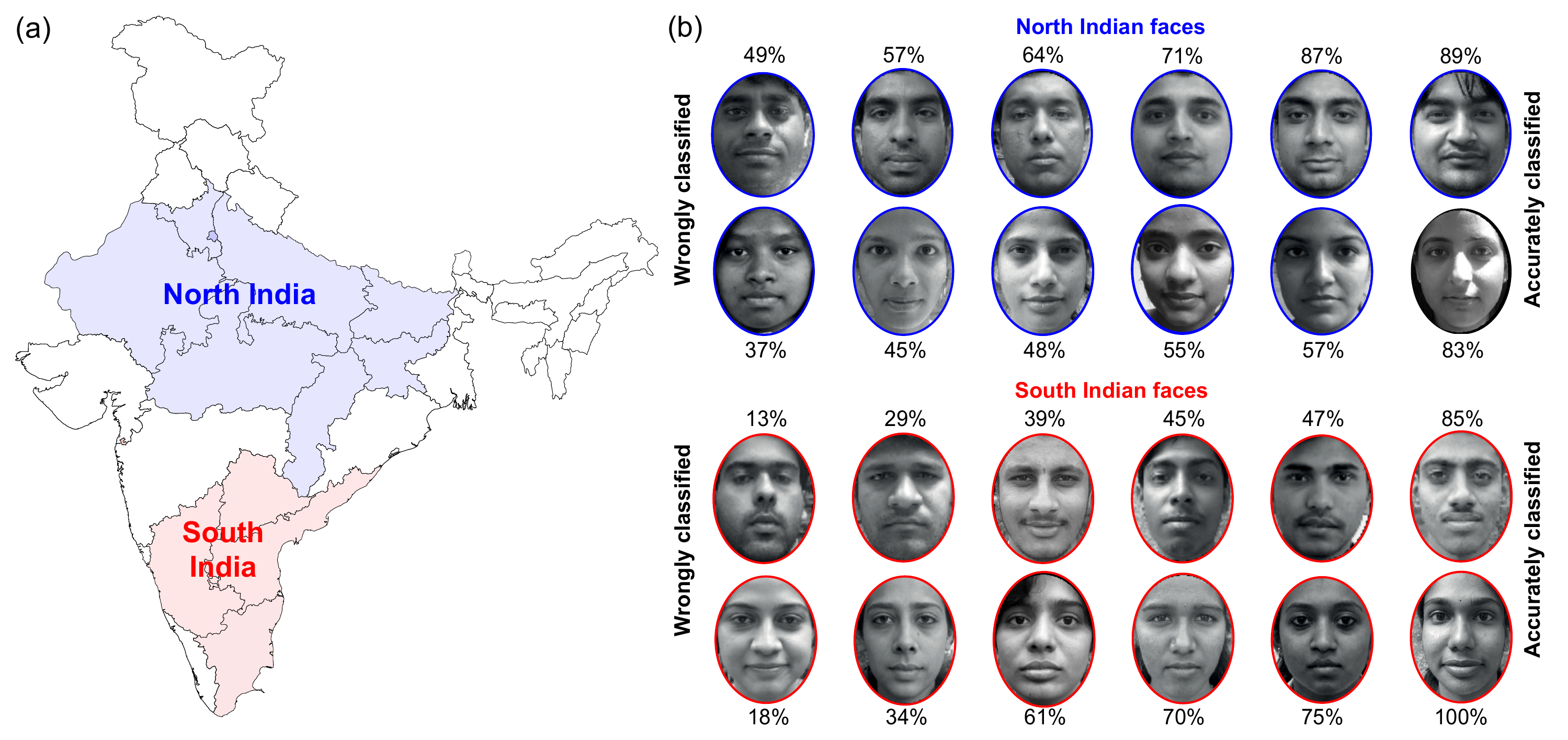}
\end{center}
\caption{\textbf{(a) Operational definition of North (\textit{blue}) and South (\textit{red}) India in our study}. We included states that are generally agreed to be part of Northern or Southern parts of India, and excluded states that have unique or distinctive cultural identities (e.g. Kerala, West Bengal, Assam). The fact that independent sets of subjects were able to easily categorize these faces correctly with high accuracy confirms the validity of our definition. \\
   \textbf{(b) Example North and South Indian faces}. North faces are shown here with a blue border and South faces with a red border, with male faces in the first row and female faces in the second row. Faces are sorted from left to right in ascending order of accuracy with which they were classified correctly across human observers. Each observer saw a given face exactly once. The text alongside each face is the percentage of human observers who correctly classified it into its respective region. In the actual experiment, subjects saw the cropped face against a black background.
   }
\label{fig:north-south-def-ex}
\end{figure*}

Here we present a fine-grained race classification problem on Indian faces that involves distinguishing between faces originating from Northern or Southern India. India contains over 12\% of the world's population with large cultural variability. Its geography can be divided roughly into Northern and Southern regions (Figure \ref{fig:north-south-def-ex}a) that have stereotyped social and cultural identities with strong regional mixing. This has resulted in stereotyped face structure, illustrated in Figure \ref{fig:north-south-def-ex}(b). Many Indians are able to classify other Indians based on the face as belonging to specific regions or even states in India, but are unable to describe the face features they are using to do so. Our goal was therefore to characterize human performance on this fine-grained race classification and elucidate the underlying features using computational models. 


\subsection{Background}
Below we review literature from both human and computer vision related to face classification. Coarse race distinctions such as Caucasian/Black have been extensively studied in humans \cite{Brooks2010}\cite{Fu2014} as well as in computers, where algorithms typically achieve 70-80\% accuracy \cite{Fu2014}. Humans can reliably classify race in the absence of salient (but potentially informative) cues such as skin colour, expressions, cosmetics, ornaments or attributes such as hair style \cite{Brooks2010}. Computational studies have also revealed that experience driven biases can contribute to asymmetries in race perception \cite{otoole2013}. 

\indent Models trained with feature extraction schemes using Local binary patterns, wavelets and Gabor filter banks have been trained to near human level performance on face recognition \cite{Fu2014} and coarse ethnicity categorisation \cite{Wang2016}. More recently deep convolutional neural networks (CNN) \cite{Krizhevsky2012} have shown impressive performance on discriminating Chinese, Korean and Japanese faces \cite{Wang2016}.

Despite the above advances, several questions remain unanswered. First, what are the underlying features used by humans? Because differences in coarse race are large, they manifest in a number of face features. This makes it difficult to identify the true subset of features used by humans. This problem is exacerbated by the fact that most successful computer algorithms ranging from local binary patterns \cite{Ojala1994} to deep neural networks \cite{Krizhevsky2012} use representations that are impossible to interpret. Second, do these algorithms behave as humans do across faces? Answering this question will require both humans and machines to exhibit systematic variations in performance across faces, which is only possible with hard classification tasks. We address both lacunae in our study. 

\subsection{Contributions}

\indent Our main contribution is to systematically characterize human performance on a fine-grained race classification task, and elucidate the underlying features using computational modeling. In particular we have shown that (1) Humans show highly systematic variations in performance on this task - sometimes even consistently misclassifying faces; (2) These variations are poorly predicted by computer vision algorithms despite explicit training on the human data; (3) Features underlying human performance can be understood by analyzing overcomplete feature representation from each face part - this analysis revealed that mouth shape was the single largest contributor towards classification; (4) We confirmed this prediction using a behavioral experiment on humans, in which we show that occluding the mouth impaired classification much more compared to occluding other parts. (5) Finally, we are making publicly available a large dataset of Indian faces with a number of ground-truth labels as well as human classification data, which will add to the relatively few datasets available for Indian faces \cite{Somanath2011,Reecha2015,Setty2013}. This dataset is available freely on github \textcolor{blue}{\textit{\href{https://github.com/harish2006/CNSIFD}{https://github.com/harish2006/CNSIFD}}}. 

\subsection{Overview}
We first describe our Indian face dataset in Section \ref{dataset}. We then describe behavioural experiments on human subjects in Section \ref{nstask}. In Section \ref{models}, we describe various feature extraction schemes and models for automatic race categorisation. In Section \ref{evaluation}, we report model and human performance. 

\section{Dataset}
\label{dataset}

Our operational definition for North and South Indians is illustrated in Figure \ref{fig:north-south-def-ex}(a). We included states that are representative of North and South India, and excluded states with unique or ambiguous identity. The fact that our participants were easily able to use this classification confirms the validity of our definition. Our face dataset has a total of 1647 Indian faces drawn from two sets of faces, as summarized in Table \ref{dataset-details}. 

\indent Set 1 consisted of 459 face images collected with informed consent from volunteers in accordance with a protocol approved by the Institutional Human Ethics Committee of the Indian Institute of Science. Volunteers were photographed in high resolution (3648 x 2736 pixels) against a neutral background. Photographs were collected primarily from volunteers who declared that they as well as both parents belong to a north Indian or south Indian state. For exploratory purposes, we also included the faces of 110 volunteers who declared themselves to be from other regions in India (e.g. Kerala, West Bengal). In addition to their race, participants were requested to report their age, height and weight as well. 

\indent Set 2 consisted of 1,188 faces selected from the internet after careful validation. Since Indian names are strongly determined by their ethnicity, we first identified a total of 128 typical first and 325 last names from each region based on independently confirming these choices with four other Indian colleagues (who were not involved in subsequent experiments). Example first names were Birender \& Payal for North, Jayamma \& Thendral for South. Example last names were Khushwaha \& Yadav for North and Reddy \& Iyer for South. We then used Google Image search APIs to search for face photographs associated with combinations of these typical first and last names. Frontal faces were detected using the CART face detector provided in Matlab's computer vision toolbox and faces in high resolution (at least 150 x 150 pixels) and for which at least 3 of 4 colleagues (same as those consulted for names) agreed upon the race label. These faces were then annotated for gender as well. 

\textit{Validation of Set 2}. Because Set 2 faces were sourced from the internet, we were concerned about the validity of the race labels. We performed several analyses to investigate this issue. First, post-hoc analysis of classification accuracy revealed that human accuracy on Set 2 (63.6\%) was similar to that on Set 1 (62.88\%) and this difference was not statistically different (p $=$ 0.51, rank-sum test comparing response correct labels of faces in the two sets). Second, we asked whether human performance was similarly consistent on the two sets. To this end, we randomly selected responses of 20 subjects from each set, and calculated the correlation between the accuracy of two halves of subjects. We obtained similar correlations for the two Sets (r = 0.73 $\pm$ 0.05 for Set 1, r = 0.71 $\pm$ 0.02 for Set 2; correlation in Set 1 $>$ Set 2 in 585 of 1000 random subsets). Finally, we asked whether classifiers trained on Set 1 and Set 2 generalized equally well to the other set. For instance it could be that the race labels of Set 2 were more noisy and therefore constituted poorer training data. To this end, we selected 400 faces from each set and trained a linear classifier based on SI features on race classification. The classifiers trained on Set 1 achieved an accuracy of 66.4\% on Set 1 and generalized to Set 2 faces with an accuracy of 55.2\%. Likewise classifier trained on Set 2 achieved an accuracy of 61\% on Set 2 and generalized to Set 1 with an accuracy of 56.5\%. Thus, classifiers trained on either set generalized equally well to the other set. In sum, the overall accuracy and consistency of human subjects as well as feature-based classification accuracy were all extremely similar on both sets. Based on these analyses we combined the race labels of both sets for all analyses reported in the paper.

\textit{Image pre-processing}. We normalised each faces by registering it to 76 facial landmarks \cite{Milborrow2014} followed by rotation and scaling such that the mid-point between the eyes coincided across faces and the vertical distance from chin to eyebrow became 250 pixels without altering the aspect ratio. We normalised the low level intensity information across faces in the dataset since some photographs were taken outdoors using histogram equalization (\textit{histeq} in Matlab\textregistered) to match the intensity distribution of all faces to a reference face in the dataset.

\begin{table}
\small
\begin{center}
\begin{tabular}{|l|r||r|r||r|r||r|}
\hline
\textbf{Face set} & \textbf{Total} & \textbf{Male} & \textbf{Female} & \textbf{North} & \textbf{South} & \textbf{Other}\\
\hline
Set 1 &  459 & 260 & 199 & 140 & 209 & 110\\ \hline
Set 2 & 1188 & 710 & 478 & 636 & 552 & 0\\ \hline
Total & 1647 & 970 & 677 & 776 & 761 & 110 \\ \hline
\end{tabular}
\end{center}
\caption{Summary of face dataset. Set 1 consisted of face photographs taken with consent from volunteers who declared their own race. Set 2 consisted of face images downloaded from the web. See Section \ref{dataset} for details.}
\label{dataset-details}
\end{table}

\section{Human behavior}
\label{nstask}

\textit{Subjects}. A total of 129 subjects (77 male, aged 18-55 years) with normal or corrected-to-normal vision performed a race classification task. All experimental procedures were in accordance to a protocol approved by the Institutional Human Ethics Committee of the Indian Institute of Science, Bangalore. 

\textit{Task}. Subjects performed a classification task consisting of several hundred trials. On each trial, a salt and pepper noise mask appeared followed by a fixation cross, each for 0.5 seconds. This was followed by a face shown for 5 seconds or until a response was made. Trials were repeated after a random number of other trials if a response was not made within 5 seconds. Subjects were instructed to indicate using a key press (N for North, S for South) whether the face shown was from North or South India. They were instructed to be fast and accurate and no feedback was given to participants about their performance. Each face was shown only once to a given subject, and a given subject saw on average 259 faces. The average number of subject responses per face was 41 for Set 1 and 28 for Set 2. 

\begin{figure}[t]
\begin{center}
   \includegraphics[trim={0 0 0 0},width=1.1\linewidth,height=0.55\linewidth]{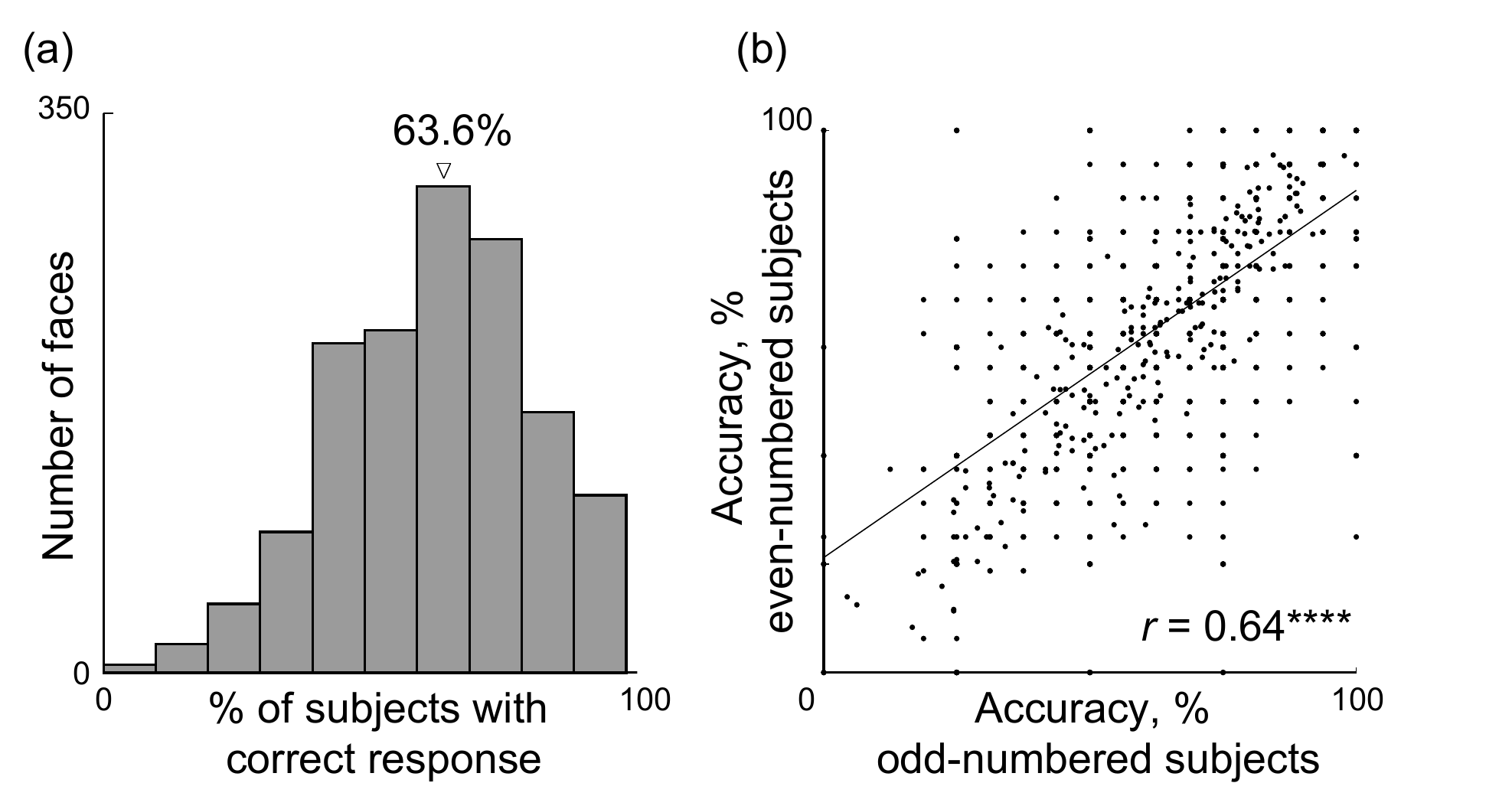}
\end{center}
\caption{\textbf{Summary of human performance}. (a) Distribution of human accuracy across faces on the race classification task. Accuracy is calculated as the fraction of participants who correctly guessed the race label. (b) Human accuracy for each face calculated from even-numbered subjects plotted against that obtained from odd-numbered subjects. The high correlation indicates that humans were highly consistent: faces that were accurately judged by one group of subjects were also accurately judged by another independent group.}
\label{fig:human-pc-distribution}
\end{figure}

\textit{Performance}. In all we obtained responses from 129 participants for 1423 faces across both sets with over 16 responses for each face. Subjects found the task challenging: the average accuracy was 63.6\%, and this performance was significantly above chance (p \textless~0.0005, sign-rank test comparing response correct labels across 1423 faces against a median of 0.5). Nonetheless there were variations in accuracy across faces, as shown in Figure \ref{fig:human-pc-distribution}(a). These variations were highly systematic as evidenced by a high correlation between the accuracy obtained from one half of subjects with that of the other half (r = 0.64, p\textless~0.0005, Figure \ref{fig:human-pc-distribution}b). 

\vspace*{-0.4pc}

\section{Computational Models}
\label{models}

\vspace*{-0.3pc}

To elucidate the features used by humans in race classification, we compared human performance with a number of computational models. We selected popular models from the computer vision literature, such as Local Binary Patterns, HOG, CNN. We also evaluated the performance of simple spatial and intensity features extracted from each face. However the problem with these models is that their underlying features are difficult to tease apart. Therefore, to elucidate the contribution of individual face parts to human performance, we evaluated the performance of a number of part-based models based on features extracted from specific face parts. 

\subsection{Local Binary Patterns (LBP) and Histogram of Oriented Gradients (HOG)}
We extracted LBP features over tiled rectangular 3 x 3, 5 x 5 and 7 x 7 patches and obtained a 1328 dimensional feature vector for each face. Our approach is similar to that in \cite{lbp2006}. 
HOG features over 8 orientations were extracted over similar patches as LBP and we obtained a dense 6723 dimensional HOG feature vector for each face. Our approach is similar in spirit to \cite{hogface2011}. 

\subsection{CNN models (CNN-A, CNN-G, CNN-F)}
The first CNN, VGG-Face~\cite{Parkhi15} is a face recognition CNN which we refer to as CNN-F. The second is a CNN trained for age classification \cite{Levi15}, which we refer to as CNN-A. The third is a CNN trained for gender classification \cite{Levi15}, which we refer to as CNN-G. CNN-A and CNN-G consist of 3 convolutional layers with 96x7x7, 256x5x5, 384x3x3 filter sizes respectively, followed by two 512-node fully connected layers and a single-node decision layer. CNN-F on the other hand is a much deeper network and has 11 convolutional layers with filter sizes varying from 64x3x3 to 512x7x7, 5 max pool layers and and 3 fully connected layers \cite{Parkhi15}. We used the penultimate 512-dimensional feature vector for each face from the CNN-A \& CNN-G network and a 4096-dimensional feature vector from CNN-F.

\subsection{Spatial and Intensity features (S, I, SI, IP, SIex, Mom)}

We also compared computational models based on spatial and intensity features extracted from each face. The spatial features were obtained by measuring a number of 2-d distances between various face parts of interest, and intensity measurements which are based on statistics of intensity in each local region of the face. We tested two approaches to evaluate these features: selective sampling and exhaustive sampling of features. 

\begin{figure}[t]
\begin{center}
   \includegraphics[width=1\linewidth]{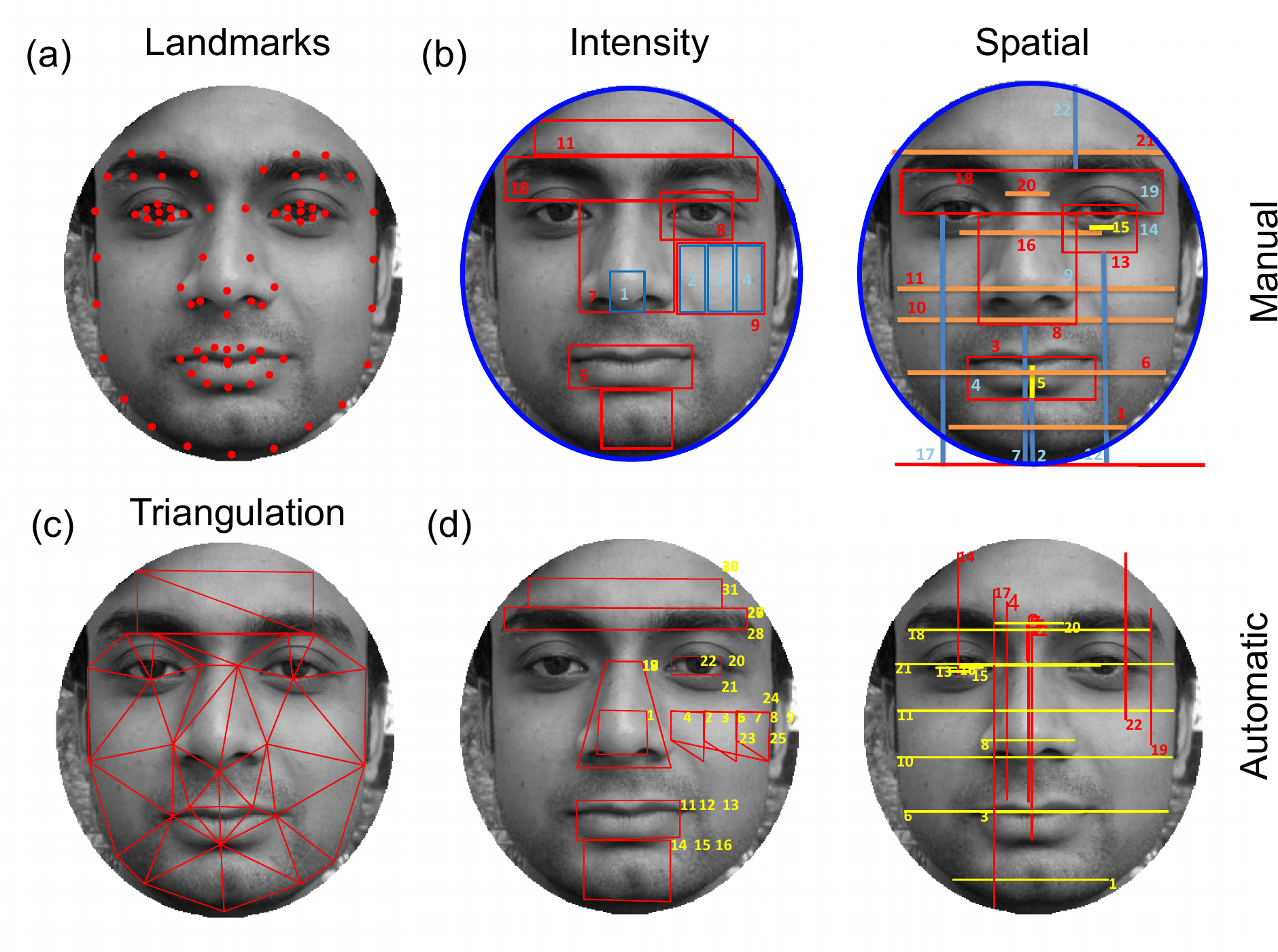}
\end{center}
   \caption{\textbf{Spatial and intensity measurements}. (a) Each face was first registered to 76 AAM landmarks using standard methods. (b) Manually defined regions for sampling intensity (left) and spatial measurements (right) outlined in blue (c) Delaunay triangulation on detected landmarks to identify all possible spatial regions on the face. (d) Measurements made from patches defined on AAM landmarks. Figure best seen in high-resolution in the digital version.}
\label{fig:spatial-intensity-aam}
\end{figure}

First, we selectively sampled spatial distances between specific landmarks and sampled intensity statistics within specific regions in the face. We started by registering an active appearance model \cite{Milborrow2014} to each face in order to identify 76 facial landmarks as illustrated in Figure~\ref{fig:spatial-intensity-aam} (a). These landmarks were then used to delineate patches and mean, minimum and maximum intensity values were recorded along with landmark based spatial features and this yielded a set of 23 spatial (S) and 31 intensity (I) measurements Figure~\ref{fig:spatial-intensity-aam}(b). 

Second, we exhaustively sampled all possible pairs of 2d distances and intensity measurements. We employed Delaunay triangulation over a restricted set of 26 landmarks from which we extracted 43 face patches Figure~\ref{fig:spatial-intensity-aam}(c), each of which covered the same region across all subjects. We extracted 325 pair-wise distances from these 26 landmarks and additionally extracted the mean, minimum and maximum intensities on all 43 patches to yield 129 intensity measurements. Together these features are referred to as SIex. To investigate the possibility that global intensity statistics may also contribute to classification, we included the first 6 moments of the pixel intensity distribution (Mom).

\subsection{Local Face Features (E, N, M, C, ENMC)}
We modeled local shape by measuring all pair-wise distances over landmarks detected on left eye (9C2 = 36 distances), right eye (9C2 = 36), nose (12C2 = 66), mouth (18C2 = 153) and face contour (15C2 = 105). 
We calculated 7C2 = 21 configural features by taking the centroid-to-centroid inter-part (IP) distances between all pairs of parts which included \textit{left eye, right eye, nose, mouth, left-contour, right-contour and chin region}. 

\subsection{Model training and cross validation}
We equated models for their degrees of freedom by reducing the  

For each model, we reduced feature dimensionality by projecting feature vectors corresponding to each face along their principal components and retained projections that explain 95\% of the variance in the data \cite{Katti2017}. Models for binary race and gender classification were trained using Linear Discriminant Analysis implemented in the Matlab\textregistered~\textit{classify} function. Regression models to predict age, height and weight were trained using regularized linear regression implemented in the Matlab\textregistered~\textit{lasso} function that optimizes the squared error subject to a sparsity constraint, as given by:

\begin{equation}
\frac{1}{n}\arrowvert y-X\beta\arrowvert _{2}^{2}+\lambda \arrowvert \beta \arrowvert_{1}
\end{equation}

where $\arrowvert . \arrowvert_{n}$ represents the $L_{n}$ norm for n = 1, 2. We use the regularization parameter $\lambda$ value that minimizes the mean square error between predicted and observed values for $y$.\\
\indent We performed 10-fold cross-validation throughout to avoid overfitting. In all cases, model performance is reported by concatenating the predictions across the 10 folds and then calculating the correlation with observed data. 

\vspace*{-0.3pc}

\section{Results from computational modeling}
\label{evaluation}
To summarize, our dataset contains 1647 Indian faces annotated with fine-grained race and gender. A fraction of faces also contained self-reported age (n=459), height (n=218) and weight (n=253) as well. In addition to this ground-truth data, we obtained race classification performance from total of 129 subjects. We tested a number of computational models for their ability to predict all these data given the face image. 

\indent Our results are organized as follows. First, we evaluated the ability of computational models in predicting fine-grained race labels. We found that while several models achieved human levels of performance, their error patterns were qualitatively different. Second, we evaluated whether models can predict human accuracy when explicitly trained on this data. This yielded improved predictions but still models were far from human performance. By comparing features derived from individual face parts, we were able to elucidate the face parts that contribute to human accuracy. Third, we investigated whether these models can predict other associated labels such as gender, age, height and weight. 

\subsection{Predicting fine-grained race labels}

	First we evaluated the ability of computational models in predicting the ground-truth race labels. The cross-validated performance of all the models is summarized in Table \ref{table:models-race}. Three models yielded equivalent accuracy for race (i.e. 63\% correct): Spatial and intensity features (SI), Histogram of gradients (HOG) and CNN-F. 
   
\begin{table}[ht]
\small
\begin{center}
\begin{tabular}{|l| |r|c|r| |r|c|r|}
\hline
 & \multicolumn{3}{|c|}{N/S classifier  accuracy} & \multicolumn{3}{|c|}{Corr with human \%} \\
\hline
\textbf{Model} & \textit{\textbf{df}} & \textbf{\%} & \textbf{R} & \textit{\textbf{df}} & \textbf{Corr} & \textbf{R}\\ \hline
\#Faces & -  & 1537  & -  & - &  1423   & -    \\ \hline
Human  & -  & 64$\pm$0\%  & -  & -  &   0.76   & -    \\ \hline
 S  & 10  & 54$\pm$0\%*  & 11  & 7  & 0.18$\pm$0.01*  & 8 \\ 
 I  & 14  & 62$\pm$0\%*  & 4  & 12  & 0.33$\pm$0.01*  & 2 \\ 
 SI  & \textbf{24}  & \textbf{63$\pm$1\%}  & \textbf{2}  & \textbf{18}  & \textbf{0.36$\pm$0.00}  & \textbf{1} \\ 
 SIex  & 56  & 57$\pm$1\%*  & 8  & 47  & 0.23$\pm$0.01*  & 4 \\ 
 Mom  & 2  & 50$\pm$0\%*  & 16  & 1  & 0.16$\pm$0.00*  & 10 \\ 
 LBP  & 172  & 54$\pm$0\%*  & 12  & 157  & 0.00$\pm$0.01*  & 17 \\ 
 HOG  & \textbf{487}  & \textbf{63$\pm$1\%}  & \textbf{1}  & 423  & 0.13$\pm$0.01*  & 15 \\ 
 CNN-A  & 124  & 60$\pm$1\%*  & 6  & 54  & 0.29$\pm$0.01*  & 3 \\ 
 CNN-G  & 52  & 59$\pm$0\%*  & 7  & 17  & 0.22$\pm$0.01*  & 5 \\ 
 CNN-F  & \textbf{735}  & \textbf{62$\pm$1\%}  & \textbf{3}  & 225  & 0.22$\pm$0.02*  & 6 \\ 
 E  & 5  & 51$\pm$1\%*  & 14  & 5  & 0.14$\pm$0.01*  & 14 \\ 
 N  & 7  & 53$\pm$0\%*  & 13  & 6  & 0.15$\pm$0.01*  & 12 \\ 
 M  & 6  & 56$\pm$0\%*  & 9  & 4  & 0.20$\pm$0.01*  & 7 \\ 
 C  & 5  & 51$\pm$1\%*  & 15  & 2  & 0.15$\pm$0.01*  & 11 \\ 
 IP  & 6  & 49$\pm$1\%*  & 17  & 3  & 0.14$\pm$0.01*  & 13 \\ 
 ENMC  & 16  & 56$\pm$0\%*  & 10  & 8  & 0.18$\pm$0.01*  & 9 \\ 
 \hline
\end{tabular}
\end{center}
\caption{\textbf{Model performance on race classification}. We trained each model on the ground-truth race labels (North vs South) and report its mean$\pm$std of 10-fold cross-validated accuracy across 100 splits. An asterisk (*) beside model performance indicates that its performance was less than the best model in more than 95 of the 100 cross-validated splits which we deemed statistically significant. \textit{Legend}: \textit{df}: degrees of freedom / number of PCs; \textit{R}: rank of each model sorted according to descending order of performance; \textit{Corr}: correlation with human accuracy across all faces. \textit{Model abbreviations}: S: spatial features; I: intensity features; SI: spatial \& intensity features; SIex: exhaustive spatial and intensity features; Mom: Moments of global pixel intensity; LBP: local binary patterns; HOG: histogram-of-gradients; CNN-A,CNN-G,CNN-F: deep networks.  E: eye; N: nose; M: mouth; C: contour; IP: inter-part distances; ENMC: eye, nose, mouth \& contour together (see text).}
\label{table:models-race}
\end{table}

\vspace*{-0.3pc}

\indent To evaluate how local features contribute to race, we calculated pairwise spatial distances between facial landmark points on each specific part of the face (eye, nose, mouth and contour). This yielded an extensive set of measurements for each part that contained a complete representation of its shape. We then asked which of these feature sets (or a combination thereof) are most informative for classification. The results are summarized in Table \ref{table:models-race}. Mouth shape was the most discriminative part for race classification, and including all other face parts did not improve performance. 


\subsection{Comparing machine predictions with human classification}
\label{human-vs-machine}

Next we wondered whether faces that were easily classified by humans would be also easy to classify for the models trained on ground-truth labels. This would indicate whether humans and computational models use similar feature representations. To this end, we computed the correlation of accuracy/error patterns between every pair of models as well as between human accuracy/errors with all models. To compare these correlations with human performance, we calculated the correlation between the average accuracy of two halves of human subjects. However this split-half correlation underestimates the true reliability of the data, since it is derived from comparing two halves of the data, rather than the full data. We therefore applied a Spearman-Brown correction on this split-half correlation which estimates the true reliability of the data, which is given as $rc = 2r/(r+1)$ where \textit{rc} is the corrected correlation and \textit{r} is the split-half correlation. 

\begin{figure*}[h!]
\begin{center}
   \includegraphics[width=0.8\linewidth,height=0.4\linewidth]{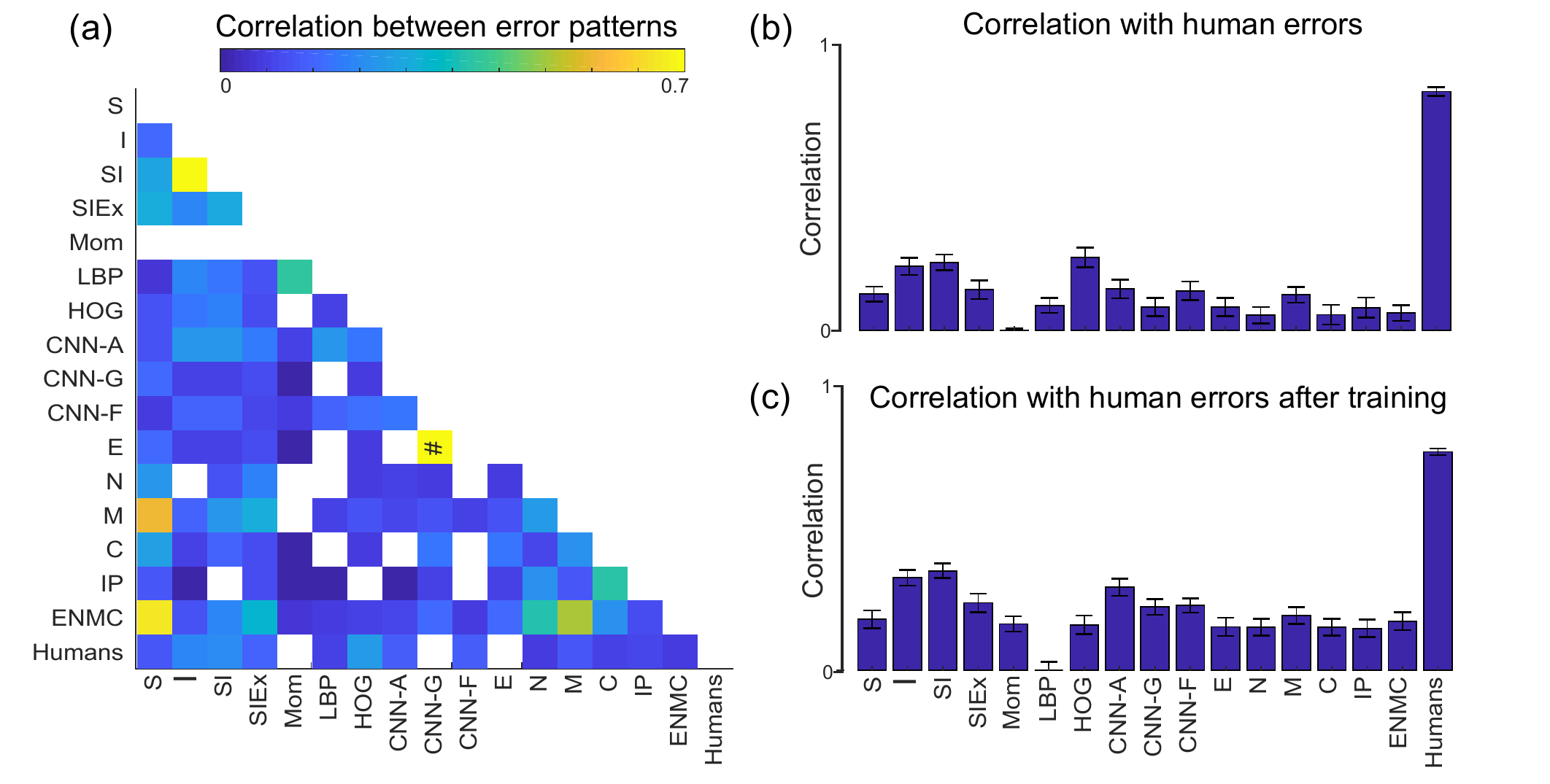}
\end{center}
   \caption{\textbf{Analysis of error patterns between models and humans}. (a) Significant pairwise correlations between the accuracy/error rate across faces for models and humans (p$<$0.05). A high artefactual correlation (r=1) between CNN-G and eye shape is marked with \textit{\#}. (b) Correlation between the accuracy of each model (trained on ground-truth labels) with human accuracy across faces. Error bars represent s.e.m calculated across 1000 iterations in which faces were sampled with replacement. The rightmost bar depicts human reliability i.e. correlation between average accuracy of one half of subjects with that of the other half of subjects. (c) Correlation between predicted and observed average human accuracy for each model. Here, models are trained to predict human accuracy.}
\label{fig:model-vs-humanpc}
\end{figure*}

In the resulting colormap shown in Figure \ref{fig:model-vs-humanpc}, models with similar error patterns across faces show high correlations. Importantly, error patterns of all models were poorly correlated with human performance (Table \ref{table:models-race} and Figure \ref{fig:model-vs-humanpc}b). This poor correlation between model and human errors could result potentially from models being trained on a mix of weak and strong race labels, or because of different feature representations. To distinguish between these possibilities, we trained models directly to predict human accuracy using regression methods. Since different features could contribute to an accurately classified North face and a South face, we trained separate models for each class and then concatenated their predictions. The resulting model performance is summarized in Figure \ref{fig:model-vs-humanpc}(c). Despite being explicitly trained on human performance, models fared poorly in predicting it. 

We conclude that human performance cannot be predicted by most computational models, which is indicative of different underlying feature representations. 

Finally, we asked whether the agreement between the responses of two different humans, was in general better than the agreement of two models using different feature types. We performed this analysis on faces from Set 1 that had a higher number of human responses than Set 2. The average correlation between correct response patterns for two human subjects (r = 0.46) was higher than the average pair-wise correlation between models trained for north/south categorization (r = 0.08) and also models trained to predict average human accuracy (r = 0.35) (p $<$ 0.0001, rank-sum test between human-human correlations and model-model correlations).

\subsection{Face part information in model representations}
\begin{figure}[h!]
\begin{center}
   \includegraphics[width=1\linewidth]{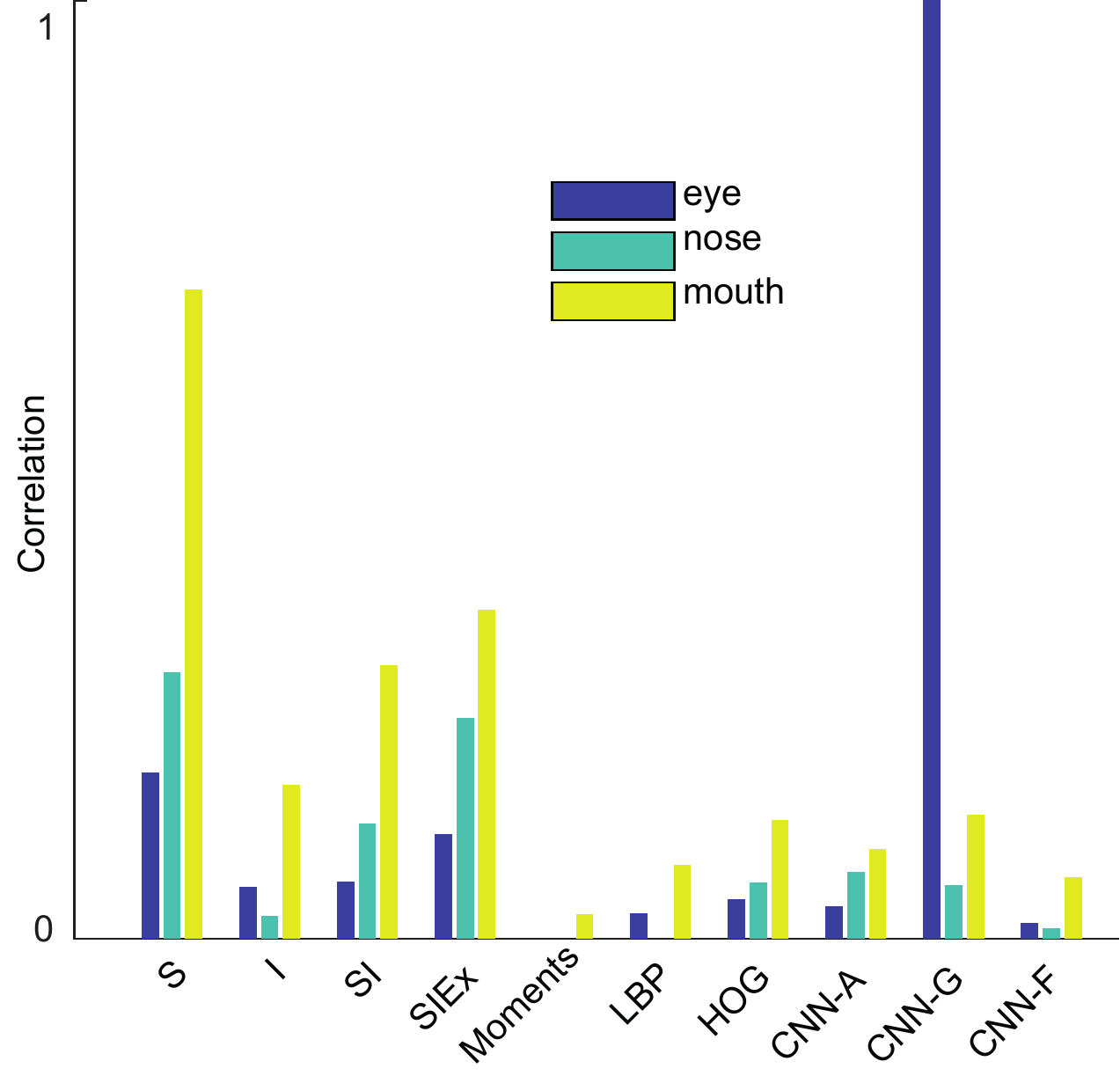}
\end{center}
   \caption{\textbf{Whole-face model predictions correlate better with mouth}. The bar plot shows the correlation between predicted race labels for models trained on whole-face information with predicted labels for models trained on eyes, nose or mouth shape alone. The sole exception is CNN-G which is particularly dependent on eye shape (see text).}
\label{fig:model-vs-enm}
\end{figure}

Since mouth shape was most informative for north vs south race labels Table. \ref{table:models-race}, we asked whether mouth shape was encoded preferentially in computational models as well? We computed pair-wise correlations between the predictions of each models trained with whole face information, with models trained with eye, nose or mouth shape alone using the method used to derive Figure. \ref{fig:model-vs-humanpc}. These are shown in Figure \ref{fig:model-vs-enm}. 

We found that local shape of the mouth is reliably encoded in the representations of all models trained with varying types of image statistics including spatial and intensity features, CNN as well as non-CNN based features. The interesting exception is the preferential encoding of eye shape information by the CNN trained for gender classification (CNN-G). This particular correlation has been reported earlier in \cite{LapAMFG17} where authors attributed it to training dataset biases and the lack of real world priors in CNNs trained from scratch with face databases and gender labels. 

\subsection{Predicting gender, age, weight and height attributes}
\label{other-labels}

Humans are adept at judging not only race but several other attributes such as gender, age, height and weight from a given face. It then becomes plausible that there is a common feature representation that can be flexibly re-weighted to learn decision boundaries for different attributes. To further investigate this, we collected all these additional attributes for as many faces as possible. All 1647 faces had gender labels, while 459 had age labels, 218 had height, 253 had weight attributes. We tested the ability of computational models in predicting all these attributes. The results are summarized in Table \ref{table:all-tasks}. 

It can be seen that all models perform gender classification much better than race classification: the best model accuracy for gender was 94\% for the HOG model, while LBP was the best model in predicting age and height, and inter-part distances were strongly correlated with weight. In general, spatial and intensity measurements (SI) were sufficient to predict gender, age, weight and height with a statistically significant correlation. Taken together the performance of these models indicates that overcomplete representations of basic spatial and intensity measurements on faces are highly informative of multiple facial attributes and re-weighting the importance of these features can give appropriate decision boundaries.

\begin{table*}
\small
\begin{tabular}{|l|c|| c|c|c||  c|c|c|| c|c|c|| c|c|c| }
\hline
  \multicolumn{1}{|c}{} & \multicolumn{1}{c||}{} & \multicolumn{3}{|c|}{Male/Female} & \multicolumn{3}{|c|}{Age} & \multicolumn{3}{|c|}{Height} & \multicolumn{3}{|c|}{Weight} \\
\hline
\textbf{Feature} & \textbf{Dims} & \textbf{df} & \textbf{\%} & \textbf{R} & \textbf{df} & \textbf{Corr} & \textbf{R}& \textbf{df} & \textbf{Corr} & \textbf{R}& \textbf{df} & \textbf{Corr} & \textbf{R}\\ 
\hline\hline
\#Faces & - & -   & 1647  & -  &   - & 459   & -   & -  & 218 &  -  & -   & 253 & -  \\
\hline
 S  & 23  & 10  & 0.68$\pm$0.00*  & 12  & 7  & 0.27$\pm$0.01*  & 2  & 7  & 0.33$\pm$0.01*  & 10  & 7  & 0.34$\pm$0.01*  & 7 \\ 
 I  & 31  & 14  & 0.75$\pm$0.00*  & 7  & 9  & 0.21$\pm$0.03*  & 9  & 10  & 0.58$\pm$0.01*  & 4  & 9  & 0.13$\pm$0.03*  & 13 \\ 
 SI  & 54  & 23  & 0.77$\pm$0.00*  & 6  & 11  & 0.26$\pm$0.02*  & 4  & 12  & 0.60$\pm$0.01*  & 3  & 12  & 0.31$\pm$0.02*  & 10 \\ 
 SIex  & 126  & 56  & 0.81$\pm$0.00*  & 3  & 2  & 0.07$\pm$0.03*  & 14  & 29  & 0.35$\pm$0.02*  & 7  & 30  & 0.41$\pm$0.02*  & 2 \\ 
 Mom  & 7  & 2  & 0.56$\pm$0.00*  & 15  & 1  & 0.03$\pm$0.02*  & 15  & 1  & 0.33$\pm$0.01*  & 9  & 1  & 0.02$\pm$0.05*  & 15 \\ 
 LBP  & 1328  & 181  & 0.56$\pm$0.00*  & 16  & \textbf{214}  & \textbf{0.36$\pm$0.02}  & \textbf{1}  & \textbf{130}  & \textbf{0.78$\pm$0.01}  & \textbf{1}  & 145  & 0.40$\pm$0.02*  & 4 \\ 
 HOG  & 6723  & \textbf{540}  & \textbf{0.94$\pm$0.00}  & \textbf{1}  & 326  & 0.26$\pm$0.02*  & 3  & 169  & 0.67$\pm$0.02*  & 2  & 194  & 0.31$\pm$0.03*  & 8 \\ 
 CNN-A  & 512  & 126  & 0.78$\pm$0.00*  & 5  & 29  & 0.21$\pm$0.04*  & 10  & 12  & 0.03$\pm$0.03*  & 16  & 37  & 0.24$\pm$0.04*  & 12 \\ 
 CNN-G  & 512  & 54  & 0.79$\pm$0.00*  & 4  & 15  & 0.18$\pm$0.02*  & 11  & 12  & 0.08$\pm$0.02*  & 15  & 12  & 0.29$\pm$0.04*  & 11 \\ 
 CNN-F  & 4096  & 764  & 0.85$\pm$0.00*  & 2  & 91  & 0.11$\pm$0.03*  & 12  & 17  & 0.09$\pm$0.02*  & 14  & 39  & 0.07$\pm$0.11*  & 14 \\ 
 E  & 72  & 5  & 0.68$\pm$0.00*  & 10  & 2  & 0.00$\pm$0.04*  & 16  & 5  & 0.24$\pm$0.02*  & 13  & 5  & 0.38$\pm$0.01*  & 5 \\ 
 N  & 66  & 7  & 0.68$\pm$0.00*  & 11  & 4  & 0.25$\pm$0.01*  & 5  & 5  & 0.32$\pm$0.02*  & 11  & 5  & 0.31$\pm$0.01*  & 9 \\ 
 M  & 153  & 6  & 0.58$\pm$0.00*  & 14  & 4  & 0.08$\pm$0.02*  & 13  & 4  & 0.26$\pm$0.02*  & 12  & 3  & 0.00$\pm$0.04*  & 16 \\ 
 C  & 105  & 5  & 0.64$\pm$0.00*  & 13  & 3  & 0.22$\pm$0.01*  & 6  & 4  & 0.36$\pm$0.01*  & 6  & 4  & 0.36$\pm$0.01*  & 6 \\ 
 IP  & 21  & 6  & 0.73$\pm$0.00*  & 8  & 5  & 0.22$\pm$0.02*  & 7  & 4  & 0.37$\pm$0.02*  & 5  & \textbf{5}  & \textbf{0.47$\pm$0.01}  & \textbf{1} \\ 
 ENMC  & 396  & 16  & 0.72$\pm$0.00*  & 9  & 8  & 0.21$\pm$0.02*  & 8  & 8  & 0.34$\pm$0.02*  & 8  & 9  & 0.40$\pm$0.01*  & 3 \\ 
\hline
\end{tabular}
\caption{\textbf{Model performance on gender, age, height and weight prediction}. To classify gender, models were trained on the face features together with gender labels. Model accuracy reported is based on 10-fold cross-validation as before. To predict age, height and weight, we projected the face features for the available faces into their principal components to account for 95\% of the variance, and then performed regularized regression the features against each attribute. \textit{Legend}: \textit{Dims}: total number of features; \textit{df}: number of principal component projections selected for classification/regression; \textit{\%}: classification accuracy; \textit{Corr}: correlation between predicted and observed attributes. \textit{R}: Rank of each model sorted in descending order of performance. Asterisks beside each model's performance (if present) indicates that its performance was lower than the best model (highlighted in bold) more than 95 of 100 cross-validated splits, which we considered statistically significant.}
\label{table:all-tasks}
\end{table*}

\vspace*{-0.3pc}

\vspace*{-0.3pc}

\section{Behavioural validation of model predictions}
	The results of Table \ref{table:models-race} show that, among individual face parts such as eyes, nose and mouth, classifiers trained on mouth features are the most accurate at fine-grained race and their performance correlates best with human performance. This in turn predicts that humans base their classification judgements on mouth shape more than on eye or nose shape. We set out to test this prediction using a behavioral experiment on humans. 
    We note that this result is by no means guaranteed simply because it was observed using computational analyses: for instance, humans might adaptively use the visible features for classification, thereby maintaining the same accuracy even when a face part is occluded. It could also be that humans use some other complex features based on other face parts that only correlate with a particular face part but can still be extracted when that part is occluded. Testing this prediction in an actual experiment is therefore critical. 

\subsection{Methods}
	In this experiment, human subjects were asked to perform fine-grained race classification on faces in which the eyes, nose or mouth were occluded in separate blocks. They also performed fine-grained race classification on unoccluded faces. Importantly, some faces were unique to each block and others were common across blocks. This approach allowed us to compare accuracy for both unique and repeatedly viewed faces across occlusion conditions. 

\subsubsection{Faces and occlusion}

From amongst the 1647 faces in the CNSIFD dataset, we chose 544 faces spanning moderate~(50\%)  to easy~(100\%) levels of difficulty with half the faces being north Indian and the other half being South Indian. We then created three occlusion conditions to evaluate the relative importance of eye, lower half of the nose and mouth shape in fine grained race discrimination. Example faces are shown in Figure \ref{figure:occlusion-expt}. The occluding band was of the same height in all three cases and we took care to avoid occluding other face parts (e.g. eyebrows while occluding eyes, or nose while occluding mouth). We then created four sets of faces corresponding to \textit{no-occlusion, eye-occluded, nose-occluded and mouth-occluded} conditions. There were a total of 217 faces in each of these conditions of which 108 faces were common to all four conditions and 109 faces were unique to that condition. We ensured during selection that the average human accuracy on north vs south categorisation on the intact versions of each set of 217 faces were comparable and around 69\% based on evaluating accuracy across the full dataset.

\subsubsection{Subjects}
We recruited 24 Indian volunteers (9 female, 25.7$\pm4.46$ years) and obtained informed consent as before. We instructed participants to indicate race labels using keypress responses (\textit{n} for north and \textit{s} for south). Each subject was presented with a unique permutation of the 4! = 24 possible permutation orders of the four occlusion blocks. Only one response was collected for each of the 217 faces shown within a condition.

\subsection{Results}
\begin{figure}[h!]
\begin{center}
   \includegraphics[trim={0 0 0 0},width=0.9\linewidth,height=0.9\linewidth]{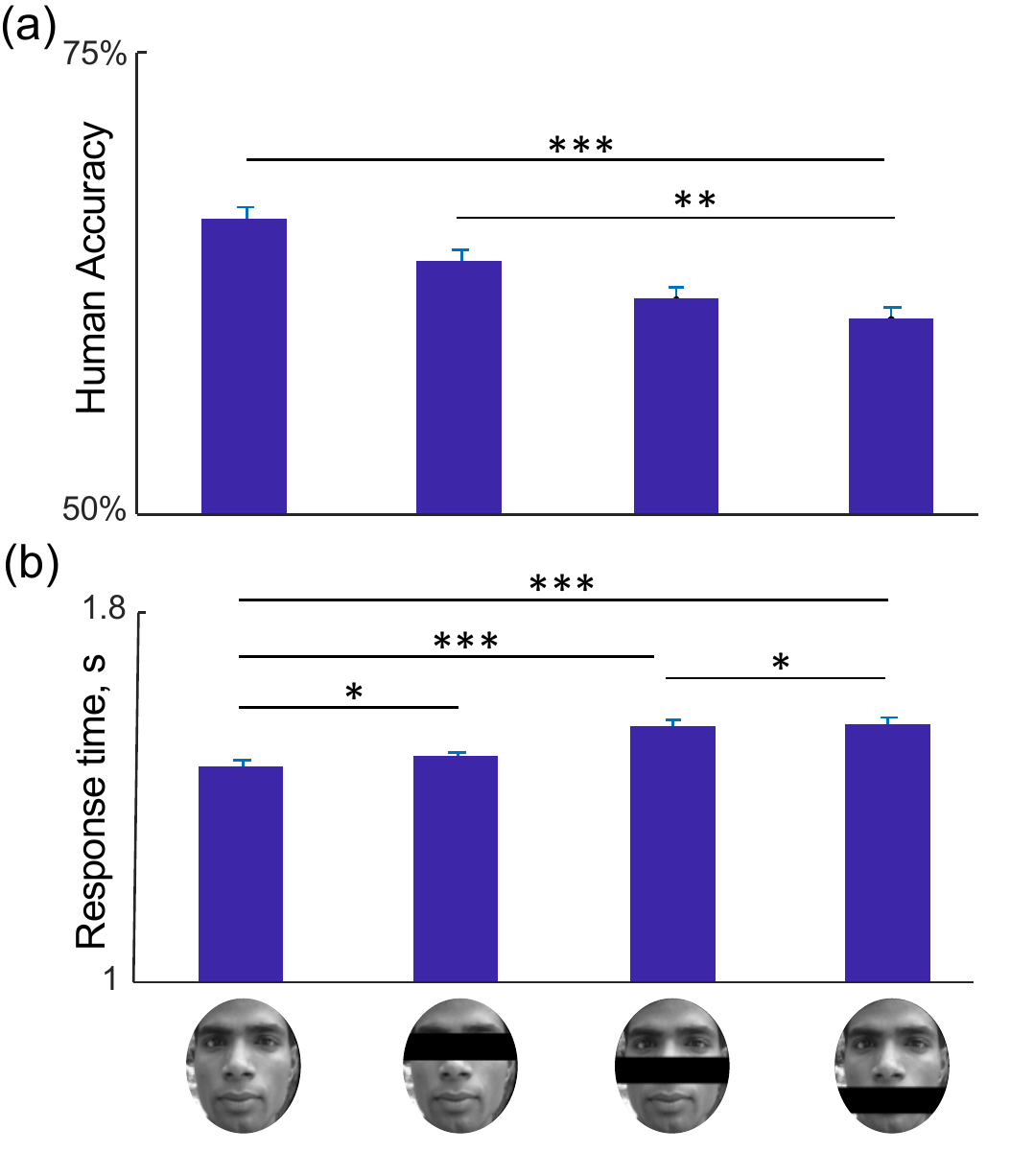}
\end{center}
\caption{\textbf{(a) Human classification accuracy and (b) Response times in each of the occlusion conditions.} Error bars indicate standard deviation about the means, \textit{*}, \textit{**} and \textit{***} denotes statistical significance of p $<$ 0.05, p $<$ 0.005 and p $<$ 0.0005 on a rank-sum test performed on binary response correct labels for all faces concatenated across subjects.}
\label{figure:occlusion-expt}
\end{figure}

	We analyzed the accuracy of subjects in each occlusion condition separately for the 108 faces common across conditions, as well as for the 109 faces unique to each condition. These data are shown in Figure \ref{figure:occlusion-expt}. Subjects were the most accurate on unoccluded faces as expected (average accuracy: 65.8\%). Importantly, classification performance was maximally impaired for the mouth-occluded (59.8\%, p$<$0.0005 compared to accuracy on unoccluded and p$<$0.005 compared to accuracy on eye-occluded faces, Wilcoxon rank sum test performed on binary response correct labels for all faces concatenated across subjects.) and nose-occluded conditions (61.1\%, p$=$0.335 compared to accuracy on mouth-occluded faces, Wilcoxon rank sum test) but not as much in the eye-occluded condition (63.6\%). Participants also faced greater difficulty in categorising mouth-occluded faces as indicated by longer response times when compared to unoccluded, eye-occluded and nose-occluded conditions (p$<$0.0005 compared to unoccluded, p$<$0.05 compared to eye-occluded or nose-occluded, Wilcoxon rank sum test performed on response times for all faces concatenated across subjects). We conclude that fine-grained race classification depends on mouth shape more than eye or nose shape, thereby validating the prediction from computational modeling. \\

\section{Discussion}
\label{discussion}

Here, we have characterized both machine and human performance on a hard race classification task. Our main finding is that while many computational models achieve human levels of performance, their error patterns are qualitatively different. This obvious discrepancy raises several interesting avenues for exploration. First, the performance gap between models and humans suggests that humans use qualitatively different features that are as yet unknown. We propose that systematically testing humans on more face recognition tasks might elucidate the underlying features. Second, the fact that humans show highly systematic variations in accuracy suggests that humans learn similar feature representations despite potentially different experiences with faces. Assuming that the kind of faces in our dataset are similar to the training examples experienced by humans, this raises the interesting question of what features humans extract from faces, and how they learn it. Both problems will be interesting for future study. 

\vspace*{-0.3pc}

\section{Acknowledgements}
We thank Pravesh Parekh, Chintan Kaur, NC Puneeth and Arpita Chand for assistance with collecting face images and human behavioral data. We are grateful to Pramod RT \& other lab members for help with curating the dataset. This work was supported by a DST CSRI post-doctoral fellowship (HK) from the Department of Science and Technology, Government of India and by an Intermediate Fellowship from the Wellcome Trust-DBT India Alliance (SPA). 
\vspace*{-0.3pc}

{\small
\bibliographystyle{model2-names}
\bibliography{egbib}

\begin{thebibliography}{19}
\expandafter\ifx\csname natexlab\endcsname\relax\def\natexlab#1{#1}\fi
\providecommand{\url}[1]{\texttt{#1}}
\providecommand{\href}[2]{#2}
\providecommand{\path}[1]{#1}
\providecommand{\DOIprefix}{doi:}
\providecommand{\ArXivprefix}{arXiv:}
\providecommand{\URLprefix}{URL: }
\providecommand{\Pubmedprefix}{pmid:}
\providecommand{\doi}[1]{\href{http://dx.doi.org/#1}{\path{#1}}}
\providecommand{\Pubmed}[1]{\href{pmid:#1}{\path{#1}}}
\providecommand{\bibinfo}[2]{#2}
\ifx\xfnm\relax \def\xfnm[#1]{\unskip,\space#1}\fi
\bibitem[{Ahonen et~al.(2006)Ahonen, Hadid and Pietikainen}]{lbp2006}
\bibinfo{author}{Ahonen, T.}, \bibinfo{author}{Hadid, A.},
  \bibinfo{author}{Pietikainen, M.}, \bibinfo{year}{2006}.
\newblock \bibinfo{title}{Face description with local binary patterns:
  Application to face recognition}.
\newblock \bibinfo{journal}{IEEE PAMI} \bibinfo{volume}{28},
  \bibinfo{pages}{2037--2041}.
\bibitem[{Brooks and Gwinn(2010)}]{Brooks2010}
\bibinfo{author}{Brooks, K.}, \bibinfo{author}{Gwinn, O.},
  \bibinfo{year}{2010}.
\newblock \bibinfo{title}{No role for lightness in the perception of black and
  white? simultaneous contrast affects perceived skin tone, but not perceived
  race}.
\newblock \bibinfo{journal}{Perception} .
\bibitem[{Duan et~al.(2010)Duan, Wang, Liu, Wu and Zhang}]{Duan2010}
\bibinfo{author}{Duan, X.}, \bibinfo{author}{Wang, C.}, \bibinfo{author}{Liu,
  Xiang-dong~Li, Z.}, \bibinfo{author}{Wu, J.}, \bibinfo{author}{Zhang, H.},
  \bibinfo{year}{2010}.
\newblock \bibinfo{title}{Ethnic features extraction and recognition of human
  faces}.
\newblock \bibinfo{journal}{2nd International Conference on Advanced Computer
  Control} .
\bibitem[{Déniz et~al.(2011)Déniz, Bueno, Salido and la~Torre}]{hogface2011}
\bibinfo{author}{Déniz, O.}, \bibinfo{author}{Bueno, G.},
  \bibinfo{author}{Salido, J.}, \bibinfo{author}{la~Torre, F.D.},
  \bibinfo{year}{2011}.
\newblock \bibinfo{title}{Face recognition using histograms of oriented
  gradientsn}.
\newblock \bibinfo{journal}{Pattern Recognition Letters} \bibinfo{volume}{32},
  \bibinfo{pages}{1598--1603}.
\bibitem[{Fu et~al.(2014)Fu, He and Hou}]{Fu2014}
\bibinfo{author}{Fu, S.}, \bibinfo{author}{He, H.}, \bibinfo{author}{Hou,
  Z.G.}, \bibinfo{year}{2014}.
\newblock \bibinfo{title}{Learning race from face: A survey}.
\newblock \bibinfo{journal}{IEEE Transactions on PAMI} \bibinfo{volume}{36},
  \bibinfo{pages}{2483--2509}.
\bibitem[{Katti et~al.(2017)Katti, Peelen and Arun}]{Katti2017}
\bibinfo{author}{Katti, H.}, \bibinfo{author}{Peelen, M.V.},
  \bibinfo{author}{Arun, S.}, \bibinfo{year}{2017}.
\newblock \bibinfo{title}{How do targets, nontargets, and scene context
  influence real-world object detection?}
\newblock \bibinfo{journal}{Attention, Perception and Psychophysics}
  \bibinfo{volume}{79}, \bibinfo{pages}{2021--36}.
\bibitem[{Krizhevsky et~al.(2012)Krizhevsky, Sutskever and
  Hinton}]{Krizhevsky2012}
\bibinfo{author}{Krizhevsky, A.}, \bibinfo{author}{Sutskever, I.},
  \bibinfo{author}{Hinton, G.E.}, \bibinfo{year}{2012}.
\newblock \bibinfo{title}{Imagenet classification with deep convolutional
  neural networks}.
\newblock \bibinfo{journal}{Neural Information Processing Systems (NIPS)} .
\bibitem[{Lapuschkin et~al.(2017)Lapuschkin, Binder, MÃ¼ller and
  Samek}]{LapAMFG17}
\bibinfo{author}{Lapuschkin, S.}, \bibinfo{author}{Binder, A.},
  \bibinfo{author}{MÃ¼ller, K.R.}, \bibinfo{author}{Samek, W.},
  \bibinfo{year}{2017}.
\newblock \bibinfo{title}{Understanding and comparing deep neural networks for
  age and gender classification}, in: \bibinfo{booktitle}{Proceedings of the
  ICCV'17 Workshop on Analysis and Modeling of Faces and Gestures (AMFG)}.
\bibitem[{Levi and Hassner(2015)}]{Levi15}
\bibinfo{author}{Levi, G.}, \bibinfo{author}{Hassner, T.},
  \bibinfo{year}{2015}.
\newblock \bibinfo{title}{Age and gender classification using convolutional
  neural networks}, in: \bibinfo{booktitle}{IEEE Conf. on Computer Vision and
  Pattern Recognition (CVPR) workshops}.
\bibitem[{Milborrow and Nicolls(2014)}]{Milborrow2014}
\bibinfo{author}{Milborrow, S.}, \bibinfo{author}{Nicolls, F.},
  \bibinfo{year}{2014}.
\newblock \bibinfo{title}{Active shape models with sift descriptors and mars}.
\newblock \bibinfo{journal}{VISAPP} .
\bibitem[{Ojala et~al.(1994)Ojala, Pietikainen and Harwood}]{Ojala1994}
\bibinfo{author}{Ojala, T.}, \bibinfo{author}{Pietikainen, M.},
  \bibinfo{author}{Harwood, D.}, \bibinfo{year}{1994}.
\newblock \bibinfo{title}{Performance evaluation of texture measures with
  classification based on kullback discrimination of distributions}, in:
  \bibinfo{booktitle}{Proceedings of 12th International Conference on Pattern
  Recognition}, pp. \bibinfo{pages}{582--585 vol.1}.
\bibitem[{O'Toole and Natu(2013)}]{otoole2013}
\bibinfo{author}{O'Toole, A.J.}, \bibinfo{author}{Natu, V.},
  \bibinfo{year}{2013}.
\newblock \bibinfo{title}{Computational perspectives on the other race effect}.
\newblock \bibinfo{journal}{Visual Cognition} \bibinfo{volume}{21},
  \bibinfo{pages}{1121--1137}.
\bibitem[{Parkhi et~al.(2015)Parkhi, Vedaldi and Zisserman}]{Parkhi15}
\bibinfo{author}{Parkhi, O.M.}, \bibinfo{author}{Vedaldi, A.},
  \bibinfo{author}{Zisserman, A.}, \bibinfo{year}{2015}.
\newblock \bibinfo{title}{Deep face recognition}, in:
  \bibinfo{booktitle}{British Machine Vision Conference}.
\bibitem[{Setty et~al.(2013)Setty, Husain, Beham, Gudavalli, Kandasamy, Vaddi,
  Hemadri, Karure, Raju, Rajan, Kumar and Jawahar}]{Setty2013}
\bibinfo{author}{Setty, S.}, \bibinfo{author}{Husain, M.},
  \bibinfo{author}{Beham, P.}, \bibinfo{author}{Gudavalli, J.},
  \bibinfo{author}{Kandasamy, M.}, \bibinfo{author}{Vaddi, R.},
  \bibinfo{author}{Hemadri, V.}, \bibinfo{author}{Karure, J.C.},
  \bibinfo{author}{Raju, R.}, \bibinfo{author}{Rajan, B.},
  \bibinfo{author}{Kumar, V.}, \bibinfo{author}{Jawahar, C.V.},
  \bibinfo{year}{2013}.
\newblock \bibinfo{title}{Indian movie face database: A benchmark for face
  recognition under wide variations}, in: \bibinfo{booktitle}{NCVPRIPG, 2013},
  pp. \bibinfo{pages}{1--5}.
\bibitem[{Sharma and Patterh(2015)}]{Reecha2015}
\bibinfo{author}{Sharma, R.}, \bibinfo{author}{Patterh, M.S.},
  \bibinfo{year}{2015}.
\newblock \bibinfo{title}{Indian face age database: A database for face
  recognition with age variation}.
\newblock \bibinfo{journal}{International Journal of Computer Applications} ,
  \bibinfo{pages}{21--28}.
\bibitem[{Somanath et~al.(2011)Somanath, Rohith and Kambhamettu}]{Somanath2011}
\bibinfo{author}{Somanath, G.}, \bibinfo{author}{Rohith, M.},
  \bibinfo{author}{Kambhamettu, C.}, \bibinfo{year}{2011}.
\newblock \bibinfo{title}{Vadana: A dense dataset for facial image analysis},
  in: \bibinfo{booktitle}{2011 IEEE International Conference on Computer Vision
  Workshops (ICCV Workshops)}, pp. \bibinfo{pages}{2175--2182}.
\bibitem[{Tariq et~al.(2009)Tariq, Hu and Huang}]{Tariq2009}
\bibinfo{author}{Tariq, U.}, \bibinfo{author}{Hu, Y.}, \bibinfo{author}{Huang,
  T.S.}, \bibinfo{year}{2009}.
\newblock \bibinfo{title}{Gender and ethnicity identification from silhouetted
  face profiles}, in: \bibinfo{booktitle}{2009 16th IEEE International
  Conference on Image Processing (ICIP)}, pp. \bibinfo{pages}{2441--2444}.
\bibitem[{Tin and Sein(2011)}]{Tin2011}
\bibinfo{author}{Tin, H.H.K.}, \bibinfo{author}{Sein, M.M.},
  \bibinfo{year}{2011}.
\newblock \bibinfo{title}{Race identification for face images}, pp.
  \bibinfo{pages}{118--120}.
\bibitem[{Wang et~al.(2016)Wang, Liao, Feng, Xu and Jiebo.}]{Wang2016}
\bibinfo{author}{Wang, Y.}, \bibinfo{author}{Liao, H.}, \bibinfo{author}{Feng,
  Y.}, \bibinfo{author}{Xu, X.}, \bibinfo{author}{Jiebo., L.},
  \bibinfo{year}{2016}.
\newblock \bibinfo{title}{Do they all look the same? deciphering chinese,
  japanese and koreans by fine-grained deep learning}.
\newblock \bibinfo{journal}{Arxiv} .

\end{thebibliography}
}

\end{document}